\documentclass[conference]{IEEEtran}
\IEEEoverridecommandlockouts

\usepackage{cite}
\usepackage{amsmath,amssymb,amsfonts}
\usepackage{algorithmic}
\usepackage{algorithm}
\usepackage{graphicx}
\usepackage{textcomp}
\usepackage{xcolor}
\usepackage{comment}
\usepackage{float}
\usepackage{booktabs}
\usepackage{multirow}
\usepackage{url}
\usepackage[hidelinks,breaklinks=true]
{hyperref}
\usepackage{array}
\usepackage{tabularx}
\usepackage{listings}
\def\BibTeX{{\rm B\kern-.05em{\sc i\kern-.025em b}\kern-.08em
    T\kern-.1667em\lower.7ex\hbox{E}\kern-.125emX}}

\begin{document}
\raggedbottom
\title{Routing End User Queries to Enterprise Databases}

\author{
\IEEEauthorblockN{Saikrishna Sudarshan}
\IEEEauthorblockA{\textit{BITS Pilani KK Birla Goa Campus}}
\IEEEauthorblockA{\textit{Goa, India}}
\IEEEauthorblockA{\textit{f20211645@goa.bits-pilani.ac.in}}
\and
\IEEEauthorblockN{Tanay Kulkarni}
\IEEEauthorblockA{\textit{BITS Pilani KK Birla Goa Campus}}
\IEEEauthorblockA{\textit{Goa, India}}
\IEEEauthorblockA{\textit{f20212615@goa.bits-pilani.ac.in}}
\and
\IEEEauthorblockN{Manasi Patwardhan}
\IEEEauthorblockA{\textit{TCS Research}}
\IEEEauthorblockA{\textit{Pune, India}}
\IEEEauthorblockA{\textit{manasi.patwardhan@tcs.com}}
\and
\IEEEauthorblockN{Lovekesh Vig}
\IEEEauthorblockA{\textit{TCS Research}}
\IEEEauthorblockA{\textit{Delhi, India}}
\IEEEauthorblockA{\textit{lovekesh.vig@tcs.com}}
\and
\IEEEauthorblockN{Ashwin Srinivasan}
\IEEEauthorblockA{\textit{Department of CSIS, BITS Pilani KK Birla Goa Campus}}
\IEEEauthorblockA{\textit{Goa, India}}
\IEEEauthorblockA{\textit{ashwin@goa.bits-pilani.ac.in}}
\and
\IEEEauthorblockN{Tanmay Tulsidas Verlekar}
\IEEEauthorblockA{\textit{Department of CSIS, BITS Pilani KK Birla Goa Campus}}
\IEEEauthorblockA{\textit{Goa, India}}
\IEEEauthorblockA{\textit{tanmayv@goa.bits-pilani.ac.in}}
}

\maketitle
\begin{abstract}
We address the task of routing natural language queries in multi-database enterprise environments. We construct realistic benchmarks by extending existing NL-to-SQL datasets. Our study shows that routing becomes increasingly challenging with larger, domain-overlapping DB repositories and ambiguous queries, motivating the need for more structured and robust reasoning-based solutions. By explicitly modelling schema coverage,  structural connectivity, and fine-grained semantic alignment, the proposed modular, reasoning-driven re-ranking strategy consistently outperforms embedding-only and direct LLM-prompting baselines across all the metrics. 
\end{abstract}


\begin{figure*}[t]
\centering
\includegraphics[width=\textwidth]{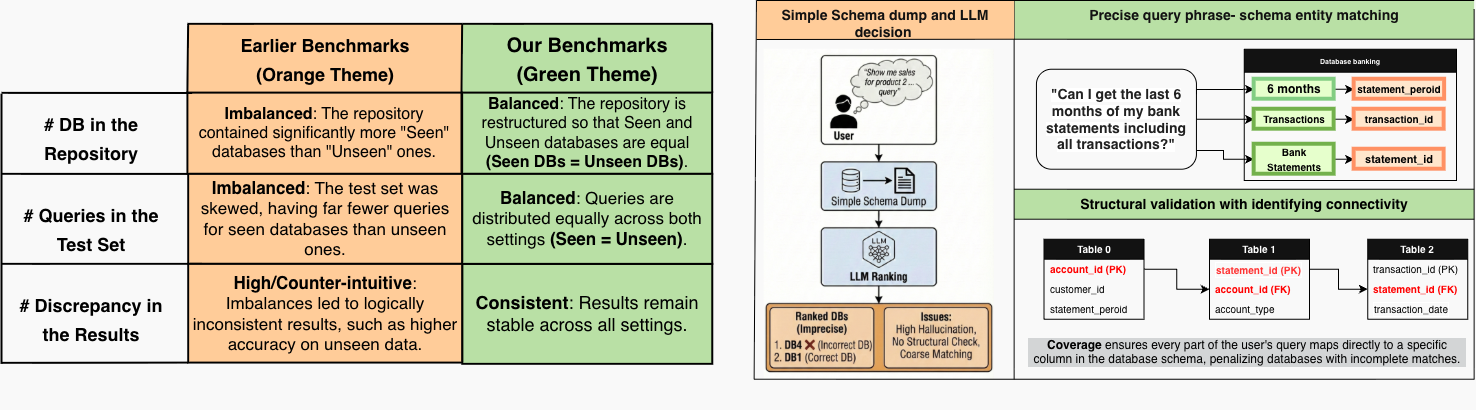}
\vspace{-5mm}
\caption{Limitations of existing benchmarks and approaches addressed by our benchmark and approach}
\vspace{-3mm}
\label{fig:dbrouting}
\end{figure*}

\section{Introduction}\label{sec:intro}

Large enterprises operate across verticals, resulting in the enterprise data being distributed across a variety of heterogeneous sources, such as knowledge graphs, relational databases, document repositories, etc. In the context of enterprise-level search, when an end-user submits a query in natural language (NL), it becomes essential to accurately route the query to the most relevant data sources that are capable of providing a correct and comprehensive response. Motivated by this practical challenge, Mandal et al.~\cite{Mandal2025DBRoutingRE} introduce a novel task of routing NL queries to appropriate data sources, specifically focusing on enterprise databases (DBs) as data sources, which encapsulate information across diverse domains.

Mandal et al.~\cite{Mandal2025DBRoutingRE} extend existing datasets developed for cross-domain NL-to-SQL semantic parsing~\cite{yu2018spider,li2024can} to construct benchmarks for the novel task of query routing, and accordingly define only in-domain and cross-domain settings. However, by preserving the original DB splits, where the test sets contain DBs only from the test split of the underlying datasets, the resulting repository size is extremely small and uneven across settings. This leads to a skewed distribution of DBs and queries in the in-domain and cross-domain scenarios, causing an imbalance in their test sets. Consequently, the cross-domain setting exhibits disproportionately higher recall and mean average precision (mAP) compared to the in-domain setting, which is counter-intuitive and highlights limitations of the benchmark in fairly evaluating routing performance under realistic data distributions.

In enterprise environments, multiple DBs frequently overlap in domain coverage, leading to shared table names, column names, and values, which significantly increases the difficulty of accurate query routing. Motivated by this and to address the above limitations, we construct a more realistic benchmark (Section~\ref{sec:dataset}). We merge all DBs from both the train and test splits of the original datasets~\cite{yu2018spider,li2024can} to form a unified DB repository and perform a 50--50\% random split of the \textbf{queries} within each DB into train and test queries. We provide both train and test splits to support future work that may require supervised training, although our proposed method itself is training-free, as DB schemas are inherently dynamic and frequent fine-tuning is impractical. In the spirit of prior work, we define two settings: (i) \textbf{In-domain}, where queries corresponding to every DB are observed during both training and testing, and the inference repository consists of \emph{all} DBs, unlike prior work where the training repository contained only train DBs and the test repository contained only test DBs; and (ii) \textbf{Cross-domain}, where queries of a subset of DBs are completely unseen during training and are introduced only at inference time, while keeping the DB repository unchanged. \textit{Note that, in either setting, a query never appears in both the training and test sets, since a 50--50 partition of queries is performed prior to defining the experimental splits.} The in-domain setting evaluates routing when query patterns of the target DBs are already observed, whereas the cross-domain setting models a realistic enterprise scenario in which DBs are changed and \textbf{new} queries must be routed to previously unseen DBs. This formulation preserves an identical repository across both settings and enables a fair and meaningful comparison.

We re-implement the techniques of~\cite{Mandal2025DBRoutingRE} as baselines on our benchmark. The inferior performance observed relative to their original results demonstrates the increased difficulty and realism of our setting. We further propose a novel training-free method that combines Schema Entity Recognition with Large Language Models (LLMs) to achieve state-of-the-art performance on this challenging benchmark. The main contributions of this work are as follows:\\

\begin{itemize}
    \item We develop more realistic and robust benchmarks for the DB routing task defined in~\cite{Mandal2025DBRoutingRE} by extending two text-to-SQL datasets, viz, Spider~\cite{yu2018spider} and BirdSQL~\cite{li2024can}.
    \item We define a new technique for the task that yields state-of-the-art results on the benchmark.
    \item We perform a detailed qualitative analysis of our results, highlighting the efficacy of our technique.
\end{itemize}

\section{Problem Statement}\label{sec:probstmt}

We have an end-user question $q$ and a repository of a set of DBs $D$, where each DB is indexed by a DB id $d_x$ with schema $S_x$ consisting of multiple tables and columns ($\{T_x\}$, $\{C_{xy}\}$), where $\{T_x\}$ is the set of tables that belong to the DB schema $S_x$ and $\{C_{xy}\}$ is the set of columns of tables $\{T_x\}$. The task is to rank the DBs in $D$ for the question $q$ based on their relevance in terms of answerability (whether the DB can provide the correct answer). We have training and test set queries $\{ q_{x_{tr}}\}$ and $\{ q_{x_{te}}\}$ for each DB $d_x$ such that $\{ q_{x_{tr}}\} \cap \{ q_{x_{te}}\} = \phi$. 


\section{Dataset Construction}\label{sec:dataset}
\begin{table}[H]
\centering
\vspace{-3mm}
\caption{\textit{Spider-Route} and \textit{BIRD-Route} Dataset Statistics}
\label{tab:dataset_summary}
\begin{tabular}{lcc}
\toprule
\textbf{Metric} & \textbf{Spider route} & \textbf{BIRD route}\\
\midrule
Total Number of DBs & 206 & 80 \\
Total Questions & 11,831 & 10,962\\
Train Questions & 5,892 & 5,461 \\
Test Questions & 5,939 & 5,501\\
\bottomrule
\end{tabular}
\end{table}

We extend two existing datasets constructed for cross-domain NL-to-SQL, viz., Spider~\cite{yu2018spider} and BIRD-SQL~\cite{li2024can}. We name the extended datasets as \textit{Spider-Route} and \textit{Bird-Route}, respectively. The dataset statistics are provided in Table~\ref{tab:dataset_summary}.
\\
\textbf{Spider-Route}\label{sec:spider-route}
A sample in the original Spider dataset~\cite{yu2018spider} consists of a DB, a NL question posed on the DB and the corresponding SQL query. Each DB is associated with comprehensive schema information, including table and column names, data types, and primary-foreign relationships, which is defined through SQL Data Definition Language (DDL) scripts. The DBs in the train and test split of the original dataset are distinct. For \textit{Spider-Route}, we combine the DBs to form a repository of 206 DBs. We formulate 50-50\% split of the questions belonging to each DB to form train and test splits for \textit{Spider-Route}, ensuring that there is no overlap between the questions. This leads us to a total number of $5,892$ training and $5,939$ test queries.
\\
\textbf{Bird-Route}\label{sec:bird-route}
The BirdSQL~\cite{li2024can} dataset is a collection of DBs spanning over a diverse set of domains such as medical, finance, education and sports. A question in the dataset is accompanied by domain knowledge, termed as `evidence', which can be used to resolve the query to SQL. A sample in the original Bird-SQL consists of a DB, a question in NL posed on the DB, question-specific evidence, and the corresponding SQL query. For each DB $d_x$, we take the union of all the question specific evidences for all the queries which belong to that DB to form DB level meta-data $m_x$. The meta-data varies from 2 to 48 sets of evidence sentences per table in the DB.
We formulate \textit{Bird-Route}, by combining the DBs of the original train (69) and dev splits (11) of BirdSQL to form the DB corpus $D$ of 80 DBs and perform 50-50\% split of questions of each DBs to form train and test splits of 5461 and 5501 queries, respectively.

\par
\section{Approach}\label{sec:meth}

In this section we elaborate on our novel technique to perform the ranking task discussed in Section~\ref{sec:probstmt}. For this technique, we use pre-trained models and do not perform any task specific fine-tuning. Thus, our method is invariant to in-domain and cross-domain settings. For each question we first compute the cosine similarity between the question embedding and the schema embeddings of the DBs in the repository. The DB schema is defined following the DDL script (Refer Table~\ref{tab:schema_output_example}) Along with the schema definition for the BIRD-Route dataset we use the meta-data in the form of column and value descriptions (discussed in Section~\ref{sec:bird-route}) for richer schema representation. The embeddings in this case are generated using the gte-Qwen2-7B-instruct~\cite{Yang2024Qwen2TR},  yielding the best results. The DBs are ranked according to the similarity scores. The top-k DBs with the highest similarity score are then selected for the re-ranking step.\\
Following~\cite{Mandal2025DBRoutingRE}, we perform re-ranking by prompting a Large Language Model (LLM) (Gemini 2.0 Flash~\cite{Comanici2025Gemini} in our case) with the top-k DB schema, their meta-data, along with the query. The prompt is reported in Table~\ref{tab:dba_instructions}. 

\begin{table}[H]
\centering
\vspace{-3mm}
\caption{Example Schema Format (DDL)}
\label{tab:schema_output_example}
\setlength{\tabcolsep}{3pt}
\footnotesize
\begin{tabular}{|p{0.44\textwidth}|}
\hline
\textbf{Database: \texttt{authors\_schema}:} 
\textbf{Table: Author}\\
\texttt{CREATE TABLE "Author" (Id INTEGER ... Name TEXT, Affiliation TEXT)}\\
Table Description: Author\\
\textbf{COLUMN 1} column\_name: Id ...\\
\textbf{COLUMN 3} column\_name: Affiliation; Column description: Organization name with which the author is affiliated.; Data format: text\\
\hline
\end{tabular}
\end{table}

\begin{table}[H]
\centering
\caption{Prompt for LLM Re-Ranking (Top-5 DBs)}
\label{tab:dba_instructions}
\setlength{\tabcolsep}{3pt}
\footnotesize
\begin{tabular}{|p{0.44\textwidth}|}
\hline
You are an expert DB Administrator tasked with evaluating and refining the relevance ranking of DBs for specific questions. You will be provided with a question and an initial ranking of the top 5 DBs deemed most relevant. Along with this you will be provided the schemas of these DBs. Your task is to analyze the question, critically assess the initial ranking, and provide a revised ranking of the top 3 DBs, where the DB at Rank 1 is *absolutely the most relevant* to answering the given question.\\
\textbf{Input:} Question [ID]: Text: [Question text]; Ranking (Top 5): Rank 1-5: [DB\_1-5]; Schemas: [Schema 1-5]\\
\textbf{Output:} Q [ID]: [DB\_1], [DB\_2], [DB\_3]\\
\textbf{Rules:} Rank 1 must be the *undisputed* best match. All 3 DBs must be distinct and from the initial ranking. Do not violate the output format or add extra text.\\
\hline
\end{tabular}
\end{table}

We observe that a direct LLM call  tends to erroneously associate query elements with the DB  elements based on lexical similarity rather than  structural intent of the query, leading to erroneous DB ranking. For example, for the query:\textit{ Find the attribute data type for the attribute named `Green'}, the DB \texttt{products\_gen\_characteristics} is incorrectly ranked higher than the ground-truth DB \texttt{product\_catalog}. This error arises because the LLM implicitly interprets the token `\textit{Green}' as a value of the \textbf{Colour} field of \texttt{products\_gen\_characteristics} DB. 
However, as specified in the query, `\textit{Green}' refers to an \textbf{attribute name} itself and not a field value of the attribute \textbf{Colour}. 

We address this limitation by proposing a novel re-ranking strategy, which allows LLMs to associate the queries more accurately with the DB schema. We decompose the problem into modular sub-tasks, allowing LLMs to focus on simplified, localised and well-scoped reasoning, rather than full end-to-end schema-query alignment. Given top-k candidate DBs and the query, we compute a `total score' and a `semantic score' for each DB for that query. We use the `total score' to re-rank the DBs for the given query. For a DB $d_x$ with schema $S_x$ and a NL query $q$, the total score indicates how well the DB $d_x$ can answer the query $q$, with $1.0$ being fully appropriate and $0.0$ being irrelevant. The `semantic score'  is used for tie-breaking when two DBs receive the same `total score'. To compute the `total score' we follow the following steps:
\\
\textbf{Step 1: Construction of Adjacency List:}
The step determines the joins that can be formed between the tables in a DB based on the schema. The LLM is prompted (prompt at ~\cite{dbroutingprompts2025}) to produce an adjacency list representing possible joins between the tables and corresponding column names, by providing the schema and table, column descriptions as the input. For example, consider a DB:  \textit{Activity}, containing the following five tables with corresponding columns: \textbf{Activity(}\texttt{activity\_id}, \texttt{activity\_name}\textbf{)}; \textbf{Participates\_in(}\texttt{student\_id}, \texttt{activity\_id}\textbf{)}; \textbf{Faculty\_Participates\_in(} \texttt{faculty\_id}, \texttt{activity\_id}\textbf{)};\textbf{ Student(}\texttt{student\_name}, \texttt{student\_id}\textbf{)};
 and \textbf{Faculty(} \texttt{faculty\_name}, \texttt{faculty\_id}\textbf{)}
The LLM is supposed to produce an adjacency list depicting possible joins amongst the tables in the schema of the form:
\{0: \{1, 2\};
1: \{0, 3\};
2: \{0, 4\};
3: \{1\};
4: \{2\}\}.
This adjacency list is used for connectivity checks in the following steps. Since this step can be performed independently of the query, it is executed just once for each DB.
\\
\textbf{Step 2: Query Phrase-Schema Entity Mapping:}
In this step, the LLM is prompted to extract the relevant query spans that can be mapped to column names and values in a DB. Specifically, the prompt ~\cite{dbroutingprompts2025}, requires the LLM to not only perform named entity recognition for the given query, but also understand the intent and semantics of the query in the context of the DB schema to identify the relevant query spans. LLM is prompted to identify not only explicit noun phrases but also verbs, adjectives, or expressions (such as average or sum), which implicitly refer to the entities in the schema. For example, for the query \textit{`What does John do?'}, the LLM is not only supposed to identify the noun \textit{`John'} to map it to an equivalent value in column \texttt{Student.student\_name} of the \textbf{Activity} DB, but also needs to identify the verb \textit{`do'} to be interpreted as referring to a person's activity and thus semantically maps to the column \texttt{Activity.activity\_name} of the DB. The query phrase identification in this context is a complex task involving both syntactic parsing and semantic associations. For a given query, this step is run for each DB, the output is a set of query phrase-to-DB schema entity mappings, where one query-phrase might be mapped to multiple DB-schema entities. For example, for the above query, the extracted phrases and entity mappings for DB {Activity} are: \texttt{John} $\rightarrow$ \texttt{Student.student\_name}; \texttt{John} $\rightarrow$ \texttt{Faculty.faculty\_name}; and
\texttt{do} $\rightarrow$ \texttt{Activity.activity\_name}.
If an identified phrase cannot be matched to any entity, it is marked as \texttt{N/A}. While in the case of a phrase with multiple mappings, all the mappings are considered valid.
\\
\textbf{Step 3: Entity Coverage and Table Connectivity:}
To assess the answerability of a query on a DB we require the query - DB pair to satisfy two criteria: Phrase coverage and Connectivity.
Phrase Coverage requires all phrases to be mapped to valid DB columns, i.e., no \texttt{N/A} mappings. A coverage score is assigned to the query DB pair according to the number of unique successfully mapped phrases.
Unmapped phrases exponentially decrease the total score to penalise missing phrases harshly. Entities with multiple mappings are counted only once. The formula for computing the coverage score = $e^{-nx}$, where 
$
x = \frac{\text{Number of NA-Mappings}}{\text{Total Number of Mappings}}, 
\quad x \in [0,1].
$
Here, $n \in [1, +\infty)$ is a hyperparameter controlling the penalty.
Small $n$ gives a softer penalty, while large $n$ penalizes \texttt{N/A} maps more harshly.
\\
Connectivity requires that all schema entities mapped from the query phrases form a connected subgraph in the database schema adjacency list. Since each phrase may admit multiple candidate mappings, the connectivity constraint is satisfied if at least one combination of mappings yields a connected component. Owing to the availability of the schema adjacency list, this can be efficiently verified using a simple graph traversal algorithm such as \emph{Breadth-First Search (BFS)} to test reachability among the selected entities. In our example, the mappings for the phrases \texttt{`John'} and \texttt{`do'} must be mutually reachable in the schema graph. There exist two valid paths that satisfy this criterion: \texttt{Table 3 $\rightarrow$ Table 1 $\rightarrow$ Table 0}, corresponding to mapping \texttt{John} as a student, and \texttt{Table 4 $\rightarrow$ Table 2 $\rightarrow$ Table 0}, corresponding to mapping \texttt{John} as a faculty member. Since at least one valid connected subgraph exists, the connectivity criterion is met, and hence the connectivity score for this query–database pair is assigned a value of \texttt{1}. If this is not the case, a score of \texttt{0} is assigned, immediately invalidating the database.

\textbf{Step 4: Total and Semantic Score Computation:}
The total score can then be computed as the product of coverage and connectivity.
We then calculate the Semantic Score as follows: For each mapped query phrase, we calculate the cosine similarity between the embeddings of the query phrase and its matched column name and information. This is done for all mapped query phrases with corresponding mapped schema entities and the semantic score is  the average of all the cosine similarity scores and is used for tie-breaking when two DBs receive the same `total score'.

Thus, the proposed method ensures that LLMs are employed judiciously for simpler subtasks where their reasoning is reliable, while structural aspects like table connectivity and coverage are handled algorithmically. This method achieves more robust and explainable DB ranking in the face of ambiguous or underspecified NL queries.

\begin{figure}[]
    \centering
    \includegraphics[width=\columnwidth, height=8cm]{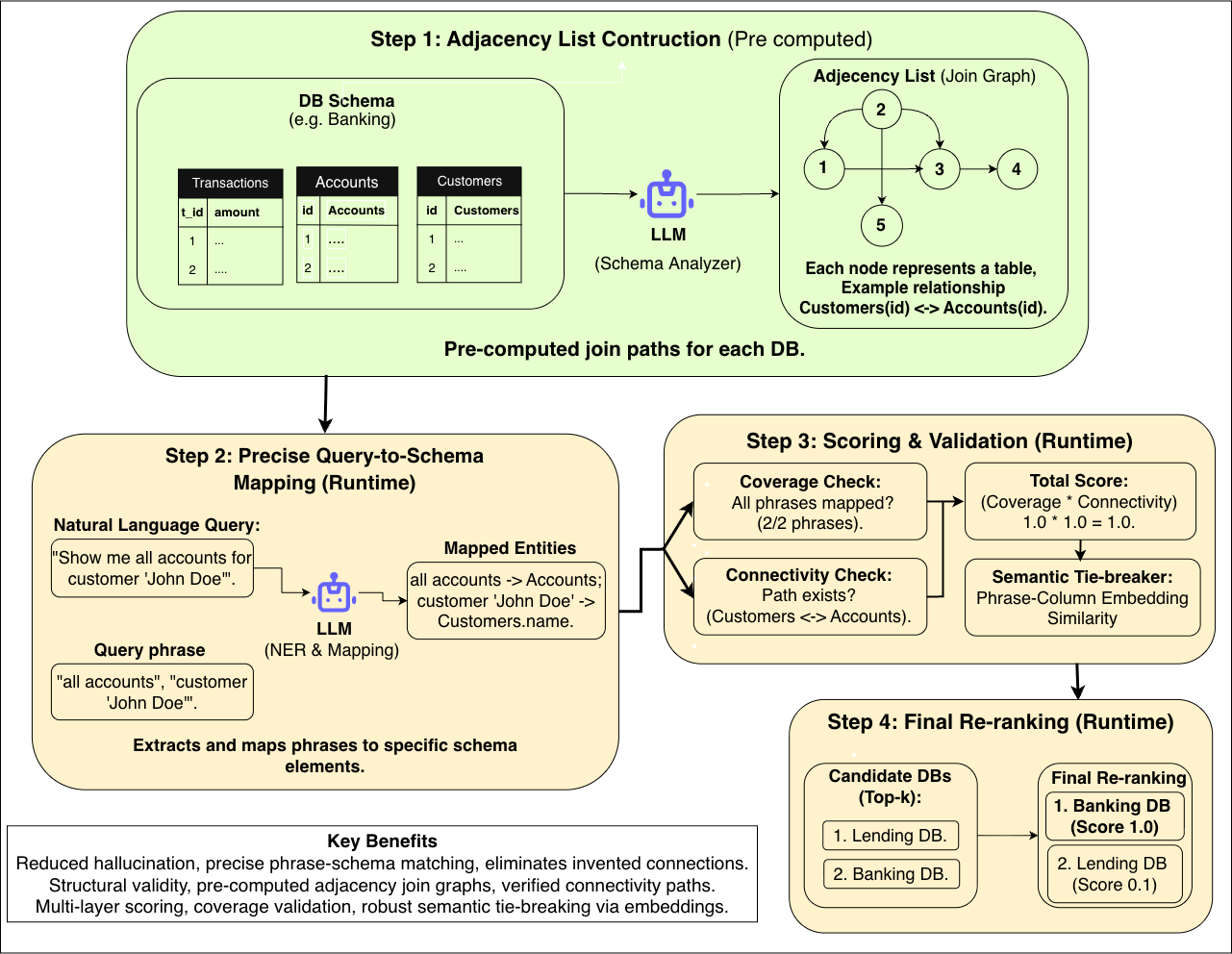}
    \label{fig:approach_overview}
    \vspace{-5mm}
    \caption{Our Approach: Modular Reasoning Re-Ranking}
    \vspace{-3mm}
\end{figure}

\begin{table*}[!h]
\centering
\caption{Results on BIRD-Route and Spider-Route Datasets}
\label{tab:combined_performance}
\resizebox{\textwidth}{!}{%
\begin{tabular}{lcccccccc}
\toprule
\textbf{Method} 
& \multicolumn{4}{c}{\textbf{BIRD-Route Dataset}} 
& \multicolumn{4}{c}{\textbf{Spider-Route Dataset}} \\
\cmidrule(lr){2-5} \cmidrule(lr){6-9}
& \textbf{R@1} & \textbf{R@2} & \textbf{R@3} & \textbf{MAP}
& \textbf{R@1} & \textbf{R@2} & \textbf{R@3} & \textbf{MAP} \\
\midrule
\textbf{\texttt{B1:}} Cosine with all-mpnet-base-v2~\cite{reimers2019sentence} embedding
& 0.4747 & 0.5733 & 0.6774 & 0.5751
& 0.4365 & 0.5191 & 0.6288 & 0.5281 \\

\textbf{\texttt{B2:}} LLM Reranking (all-mpnet-base-v2 + Llama 3~\cite{Touvron2023LLaMAOA})
& 0.5429 & 0.5971 & 0.7286 & 0.6229
& 0.4941 & 0.6325 & 0.6512 & 0.5926 \\

\textbf{\texttt{B3:}} Cosine with gte-Qwen2-7B-instruct ~\cite{Yang2024Qwen2TR} embedding
& 0.6249 & 0.7821 & 0.8344 & 0.7471
& 0.6033 & 0.6835 & 0.7471 & 0.6780 \\

\textbf{\texttt{B4:}} LLM Reranking with Gemini~\cite{Comanici2025Gemini}
& 0.7120 & 0.8060 & 0.8490 & 0.7890
& 0.6799 & 0.7166 & 0.7613 & 0.7193 \\

\textbf{\texttt{Ours:}} LLM Re-ranking with Modular Reasoning
& \textbf{0.7962} & \textbf{0.8176} & \textbf{0.8491} & \textbf{0.8210}
& \textbf{0.7865} & \textbf{0.7942} & \textbf{0.8086} & \textbf{0.7964} \\

\bottomrule
\end{tabular}%
}
\vspace{2mm}
\begin{center}
\textbf{Note:} \textbf{\texttt{B:}} Baseline. Max Possible R@1: BIRD = 82.69\%, Spider = 87.00\% (limited by top-5, hence max R@1 post-reranking = R@5 pre-reranking).
\end{center}
\end{table*}

\section{Results and Discussion}
\textbf{Metric:}
For each question, we calculate the recall@1 (R1), recall@3 (R3) and mean average precision (mAP). For a question, Recall@k is 1 if the DB ranked in the top-k by an approach is the ground truth DB and is 0 otherwise. 
\\
\textbf{Baselines:}
We use the following approaches defined in~\cite{Mandal2025DBRoutingRE} as our baseline approaches: (i) all-mpnet-base-v2~\cite{reimers2019sentence} embeddings (\textbf{\texttt{B1}}): We embed the question as well as DB schema usig all-mpnet-base-v2~\cite{reimers2019sentence} model. For a given question we find cosine similarity of its embeddings with the embeddings of all DBs and use this similarity as a scoring function to rank all the DBs for that question. (ii) LLM Re-ranking (\textbf{\texttt{B2}}): We use the top-k DBs identified by the embedding based method and use the prompt defined in~\cite{Mandal2025DBRoutingRE} to employ LLMS (Gemini~\cite{Comanici2025Gemini}) re-rank those DBs for the given question. Consequently, the recall@1 for this approach is inherently limited by the recall@5 of (\textbf{\texttt{B1}}). We also use gte-Qwen2-7B-instruct~\cite{Yang2024Qwen2TR} embedding model to get results of (i) (\textbf{\texttt{B3}}) and (ii) (\textbf{\texttt{B4}}) above, which serves as a better embedding model for us. We replace the LLM based re-ranking method with our re-ranking method and use top-k DBs identified by the gte-Qwen2-7B-instruct embedding based method for re-ranking. For our experimentation we use the value of k to be 5.

The results on \textit{Spider-Route} and \textit{Bird-Route} are discussed in Table~\ref{tab:combined_performance}. We also answer the following research questions:
\\
\textbf{Q1: Is the newly created benchmark more robust than the prior benchmark for the DB Routing task?}
As discussed in section~\ref{sec:dataset}, as opposed to the approach in Mandal et al.~\cite{Mandal2025DBRoutingRE}, we re-create the dataset leading to consistent number of DBs in the repository across in-domain and cross-domain settings.
Overall, the number of DBs in the repository is larger in our benchmark as compared to the prior one, making the task harder and more realistic. This is also demonstrated by the performance on the prior and our benchmark with the baseline approach \textbf{\texttt{B1}}. \textbf{\texttt{B1}} leads to recall@1 of 44.09 on \textit{Spider-Route} and 80.21 on the \textit{BIRD-Route} datasets of~\cite{Mandal2025DBRoutingRE}, respectively. Whereas, recall@1 of the same method \textbf{\texttt{B1}} on our \textit{Spider-Route} and \textit{BIRD-Route} datasets is 47.47  and 43.65 , respectively (Table~\ref{tab:combined_performance}). 
The proposed organisation for the \textit{Spider-Route} and \textit{BIRD-Route} dataset in this work, having more number of DBs in the repository creates a harder setting to differentiate amongst the DBs for a query, especially the ones which are closer with respect to the semantics of its schema.

\textbf{Q2: Does similarity in domains of the data-sources affect the performance of the DB-Routing ?}
We first encode the DDL schema definitions of the DBs using the gte-Qwen2-7B-instruct \cite{Yang2024Qwen2TR} embedding and then cluster using a constrained K-means algorithm grouping semantically similar DBs. A size constraint is enforced, ensuring even distribution of DBs across clusters, such that each cluster belongs to a domain. The DBs of \textit{Spider-Route} result in 10 clusters containing 20 to 23 DBs each, whereas DBs of \textit{Bird-Route} result in 5 clusters containing 15 to 17 DBs each.
We observe that for 82\% of questions for \textit{Spider-Route} and 76\% of questions of \textit{Bird-Route}, the top-5 ranked DBs belong to the same cluster and thus same domain. Thus, for the DB-Routing task DB selection is hardest within the same domain.

To quantify in-cluster confusion in our semantic embedding-based approach, we analyze all cases where the correct DB is \textbf{not} ranked at top-1. Among these errors, 86.2\% and 75.0\% of questions in \textit{Spider-Route} and \textit{Bird-Route} have the top-1 DB from the same domain cluster as the ground truth. Further, 82.3\% and 76.2\% of questions in \textit{Spider-Route} and \textit{Bird-Route} contain multiple same-domain DBs within the top-5 results. These statistics reveal that semantic embedding errors are predominantly concentrated within the similar semantic domains rather than across unrelated domains, leading to higher intra-cluster error rate. For example, in the \textit{Spider-Route} dataset, when routing fails for book-related queries, the system typically confuses between \texttt{book\_2}, \texttt{book\_press}, and \texttt{book\_review} DBs rather than incorrectly selecting unrelated DBs from sports or healthcare domains. With higher intra-cluster error rate, the \textit{Spider-Route} dataset demonstrates more cluster cohesion (tighter within-domain grouping) and relatively clear decoupling (inter cluster domain separation) when compared to the \textit{Bird-Route}. This highlights the need for a better re-ranking approach which improves within-domain discrimination capabilities.
\\
\textbf{Q3: Does our approach effectively address the challenge of discriminating between highly similar DBs in the DB Routing task?}
The proposed method outperforms the direct LLM-based re-ranking method for the DB routing task (Table ~\ref{tab:combined_performance}). For the \textit{Spider-Route}, LLM-based re-ranking and the proposed method have similar Recall@3. The results suggest that the direct LLM-based re-ranking is capable of identifying relevant DBs but cannot select the most appropriate DB for a given query. Whereas, our proposed method can distinguish between most similar DBs to identify the right DB for the query.
This is due to several key factors that enable more accurate DB re-ranking, particularly in scenarios involving multiple valid candidate DBs, precise entity matching requirements, and complex join operations. The improvement can be attributed to the three main reasons: 
\\
(i) \textit{\textbf{Precise query phrase-schema entity matching:}} The second step of our approach takes into consideration the coverage of query phrases by the schema for scoring for the re-ranking task. This ensures a more fine-grained similarity score as opposed to a coarse grained similarity check between the complete query and the schema. For example, consider the query discussed earlier:\textit{ Find the attribute data type for the attribute named `Green'}, the proposed method correctly identifies \texttt{product\_catalog} as the DB with the total score of 1.0 due to its proper attribute structure. The other top ranked DBs \texttt{products\_gen\_characteristics} and \texttt{phone\_1} receive the total score of 0.188875 and 0, respectively. In contrast to this the baseline LLM re-ranker incorrectly gives the following top-3 ranks: 1.\texttt{products\_gen\_characteristics}, 2.\texttt{product\_catalog} and 3.\texttt{phone\_1}, ranking an inappropriate DB first. 
\\
\textit{\textbf{Structural validation with identifying connectivity between the schema entities mapped to query phrases:}} Steps 1 and 3 allow taking into consideration the connectivity of the schema entities mapped to the query phrases for the re-ranking score computation. The DB with perfect connectivity score provides structural validation for the query. It ensures that the downstream retrieval of the data from the DB would be possible by writing equivalent SQL for the given query, allowing joining the tables involved in the query. For example, for the query\textit{ `What is the transmitter of the radio with the largest ERP\_kW?'}, our method correctly identifies \texttt{tv\_shows} as the top DB with a partial match of score 0.8265. Whereas for other top ranked DBs ( \texttt{program\_share}, \texttt{bbc\_channels}, \texttt{geo}), we observe that the DB entities mapped to the query phrases are distributed across tables that cannot be joined, leading to a connectivity score of 0. Since \texttt{total score =  coverage x connectivity} these DBs  are ranked lower than the correct DB. Whereas, the baseline LLM re-ranker incorrectly performs the following top-3 re-ranking: \texttt{geo}, \texttt{bbc\_channels}, \texttt{program\_share}, selecting DBs with no joining paths between the mapped DB entities.
\\
\textit{\textbf{Effective handling of multiple valid candidates through semantic tie-breaking:}} This step allows fine-grained similarity computations between the mapped query phrases and schema entities, allowing distinguishing between semantically similar DB schemas with effective tie-breaking mechanism. In contrast, the LLM reranker fails to make such fine-grained comparisons between top-k DBs. For example, for the query \textit{`How many books are there?'} with \texttt{book\_2} being the gold DB on which the query is posed. 
The baseline LLM re-ranker provides following DBs as the top-3 re-ranked list: \texttt{book\_review}, \texttt{book\_press}, \texttt{book\_2}. However, our method returns the following as the re-ranked list: \texttt{book\_2}, \texttt{book\_review}, \texttt{book\_press}, with total score of all the DBs being 1, leading to the case of a tie. This indicates that all the above DBs are capable of addressing this query. However, our method results in the \texttt{book\_2} DB obtaining the highest semantic score of \textbf{0.688}. As mentioned prior, the semantic score is computed based on the embedding similarity between the query phrase and the corresponding DB entity along with its originating table. In this case, the query term \emph{book} aligns most strongly with \texttt{books.BookID} from the \texttt{books} table in the \texttt{book\_2} DB, thereby surpassing alternative but weaker alignments such as \texttt{books.ISBN} from the \texttt{book\_1} DB and \texttt{book\_press.bookID} from the \texttt{book\_press} DB.
\\
\textbf{\textit{Error Analysis:}} Though our approach demonstrates improvement in the scores still there are 20\% and 21\% queries in \textit{Spider-Route} and \textit{Bird-Route} (Table~\ref{tab:combined_performance}), which do not have 100\% Recall@1. We perform error analysis of these queries. 11\% and 16\% of these queries, respectively, do not have the correct DBs in the top-5 results of the embedding based ranking method \texttt{B3}. For such queries we add the correct DB in top-5 and run our re-ranking method. We observe that for 98\% and 96\% of such queries for \textit{Spider-Route} and \textit{Bird-Route}, even though the DB schema has the lower cosine similarity score with the query embeddings, the proposed method is able to rank it as the top-1 for the query. This indicates that if the right DB is available for re-ranking, our method can accurately identify that DB to be the most relevant one. We perform one more experiment on a subset of 500 erroneous queries of \textit{Spider-Route} and \textit{Bird-Route}. We use Gemini 2.5 \cite{Comanici2025Gemini} as opposed to Gemini 2.0 as the base LLM and use top-20 candidate DBs from the rankings provided by \texttt{B3}. On these 500 erroneous questions with our method we get recall@1 as 92\% and 89\% on \textit{Spider-Route} and \textit{Bird-Route}. This demonstrates that use of better LLM and choosing higher k for top-k ranked DBs, our approach leads us to substantial improvements in the results. We performed this study on a subset due to much higher cost of Gemini 2.5.

For the remaining 9\% and 5\% failed queries of \textit{Spider-Route} and \textit{Bird-Route}, it is observed that the queries are ambiguous. We find that the DB selected by our method as the top-1 can also be a possible DB which can answer the query. As our benchmark uses queries, originally designed for DB specific NL to SQL task, the queries have inherent ambiguities with regards to answerability with respect to multiple DBs. They can be answered with multiple DBs. For example, for the query `How many customers in state of CA?' the ground truth DB is \texttt{store\_1}, whereas our method returns \texttt{loan\_1} as the top-1, and the ground truth DB is ranked $3^{rd}$. Both DBs contain customer information with state and location attributes and can legitimately answer this query.

This demonstrates that our re-ranking approach identifies correct DBs for $\sim$80\% of the queries. The remaining 20\% queries either (i) can be resolved by using higher k value to choose top-k ranked DBs for re-ranking and by using more reliable base LLM, or (ii) are ambiguous, where the top ranked DB by our method is also a valid DB to answer the query along with the ground truth DB. Thus, our method provides a good solution for the DB-Routing task. 
\\
\textbf{Q4: Does domain-specific knowledge facilitate improving the performance of the DB-Routing task?}
Along with the DB schema, for the \textit{Bird-Route} dataset the input for the DB Routing contains the meta-data. 
We perform an experiment without inclusion of this meta-data and observe that there is 5\% decrease in Recall@1 and 7\% decrease in Recall@3. The domain-specific knowledge in the form of meta-data contains information about the purpose of tables, the meaning of ambiguous or polysemous columns. Thus, domain-specific knowledge  provides additional semantic context to route queries in scenarios where DDL script of the schema alone cannot provide sufficient context to resolve them, leading to better performance.

\section{Conclusion}

This work introduces a principled framework for routing natural language queries to enterprise databases by combining more realistic benchmarks with a modular, reasoning-driven re-ranking strategy. By explicitly modeling schema coverage, structural connectivity, and fine-grained semantic alignment, the proposed method consistently outperforms embedding-only and direct LLM-based baselines, especially in challenging, closely similar database scenarios. These results highlight that decomposing query routing into interpretable and verifiable sub-tasks is crucial for building accurate, scalable, and reliable enterprise DB routing systems.

\end{document}